# Unveiling Public Opinion: A Study of Sentiment Analysis Using LSTM and Traditional Models

Atiq Ur Rehman
Graduate Student, Department of Economics and Decision Sciences, University of South Dakota, USA
atiq.ur.rehman.shafeeq@gmail.com, atiqur.rehman@usd.edu

***Abstract*—In this age of social media, sites like Twitter have become meeting places for people to share their views and feelings on a wide range of issues and current events as they unfold in real time. Sentiment analysis, a critical application of NLP, has become indispensable due to the massive influx of user-generated content, enabling the extraction of meaningful insights from the opinions and emotions expressed in textual data. Sentiment analysis on Twitter employs sophisticated computational techniques to categorize tweets into positive, negative, or neutral sentiments. This method not only examines individual expressions but also analyzes vast databases related to specific subjects or events. By spotting these emotions, machine learning models help improve public opinion interpretation and trend forecasting. This paper examines the effectiveness of various machine learning and deep learning approaches. Designed for this use, the system evaluates logistic regression, random forest, naïve bayes, gradient boosting, and LSTM networks, among other algorithms applied in sentiment classification. This work identifies the optimal sentiment analysis model using a Kaggle Twitter dataset that has been preprocessed through tokenization, lemmatization, and stopword elimination. Emphasizing the better performance of the LSTM approach, the model attained a training accuracy of 90.98%, a testing accuracy of 80.00%, and a micro-average ROC- AUC score of 0.92. These results show that the model outperforms conventional machine learning techniques in capturing contextual and sequential textual aspects.***

## I. Introduction

Sentiment analysis is a crucial branch of natural language processing (NLP) that focuses on extracting and interpreting subjective opinions from text-based data [1]. Understanding sentiment offers valuable insights into consumer behavior, societal trends, and market dynamics, driven by the increasing reliance on digital platforms for communication [2]. Modern sentiment analysis uses multimodal methods to analyze textual, auditory, and visual data. This integration enhances analysis by providing a deeper understanding of human emotions and sentiments [3]. Social media makes more public opinions and sentiments available than ever. Sentiment analysis is crucial for comprehending public opinion in various fields, including business and politics [6]. Sentiment analysis is also used in health, social policy, e-commerce, and digital humanities. Privacy concerns and dataset biases must be addressed in each domain to deploy sentiment analysis tools ethically and effectively [4].

As shown in Figure 1, machine learning classifiers like logistic regression, naive Bayes, and support vector machines rely on statistical models to predict sentiment [6]. On the other hand, deep learning approaches, such as convolutional neural networks (CNNs) and long short-term memory (LSTM) networks, use artificial neural networks to analyze complex patterns in text for sentiment prediction [7]. Additionally, ensemble learning techniques combine various classifiers to improve the accuracy and effectiveness of sentiment analysis [5].

Analyzing extensive social media comments to categorize products or services can be time-consuming [8]. Twitter has emerged as a widely used platform for expressing strong emotions, making it an ideal source for sentiment-related data, where users often share and discuss their opinions [9]. While machine learning algorithms can perform tasks quickly, they usually fall short in terms of performance and accuracy [10]. Textual sentiment classification is where deep learning approaches outperform machine learning [11]. Recent studies [12], [13] show that deep learning for sentiment analysis transforms text data analysis. Neural networks that identify complex patterns and relationships outperform traditional machine learning methods [14], giving them superior accuracy and performance [15], [11]. Adding text, images, and audio to sentiment analysis applications enriches insights [15]. Businesses can dynamically refine strategies by monitoring and responding to customer sentiment with real-time processing. These models improve iteratively through training and data exposure, enhancing accuracy and adaptability [13]. These advantages collectively make deep learning and Twitter data the best choice for this research, as they can accurately, adaptively, and efficiently analyze complex and large-scale sentiment data, especially Twitter data that requires real-time and contextual sentiment analysis. Machines do not understand words and letters. For a machine to understand text data, it must represent it numerically [9]. The study aims to use advanced deep learning methods to develop and deploy a high-accuracy LSTM-based sentiment analysis model. It also evaluates machine learning models to identify the most effective method for performing real-time sentiment analysis in complex and context-dependent situations. This research relies on pre-selected tweets that were automatically analyzed for emoticons, rather than being manually labeled. The dataset consists of ten fields in a CSV file, including the tweet's polarity (0 = negative, 2 = neutral, and 4 = positive), TextID, date and time, country, population, and more. Twitter sentiment analysis provides businesses with valuable insights into customer feedback, enabling them to improve their products and services [20].

## II. Literature review

Li (2024) [21] explores the most recent advancements in e-commerce sentiment analysis, emphasizing RNNs and CNNs, deep learning algorithms that aid in understanding consumer reviews and emotion-driven marketing more effectively. The work demonstrates how advanced preprocessing and feature extraction can enhance sentiment classification accuracy, addressing issues such as ambiguity, sarcasm, and contextual interpretation. It highlights how sentiment analysis and emotional marketing complement each other, using case studies to illustrate their practical benefits. The results underscore the importance of understanding consumer opinions to enhance marketing plans and foster company growth.

G. Srivastav, S. Kant, D. Srivastava, N. Sharma and Y.-C. Hu, (2024) [22] Examine the enhanced EdBERT

model for sentiment analysis in these contexts to ascertain student opinions of online learning. The primary areas of focus in this domain-specific EdBERT adaptation are academic language analysis, course content feedback, instructor performance, and platform usability. Employing EdBERT with neural network layers and advanced preprocessing, this model can capture intricate emotions, including happiness and sadness. This approach highlights the importance of deep learning models, particularly in domains. Based on the findings, EdBERT could be a fantastic tool for guiding improvements and providing comments for online learning systems.

Geethanjali & VALARMATHI, (2024) [23] This study presents a hybrid deep learning model that combines IChOA-CNN with LSTM for social media sentiment analysis, emphasizing multilingual and multimodal data. To handle images, videos, and text, the model utilizes convolutional layers and hierarchical attention mechanisms to capture subtle emotional signals. LSTM networks control temporal dependencies to help consecutive posts better. By training the model on a multilingual dataset and analyzing how culture and language influence sentiment expression, the paper demonstrates its flexibility.

M. M. Rahman, A. I. Shiplu, Y. Watanobe, and M. A. Alam, (2024) [31] This study addresses sentiment analysis challenges, including lexical diversity, long dependencies, unknown symbols, and imbalanced datasets, using several models. The hybrid RoBERTa-BiLSTM model uses RoBERTa's robust word embeddings and BiLSTM's contextual semantic learning for long texts to overcome these challenges. On the IMDb, Twitter US Airline, and Sentiment140 datasets, it beats state-of-the-art models with accuracies of 92.36%, 80.74%, and 82.25%, respectively, and F1-scores of 92.36%, 80.74%, and 82.25%. According to the model, sentiment analysis gains from combining the Transformer and sequential approaches.

Chanda & Pal (2024) [24] The study integrates deep learning with machine learning to detect hate speech in code-mixed social media content. It addresses the complexities of multilingual and dialect-rich text by employing preprocessing techniques such as tokenization, normalization, and handling unique characters to prepare the data for analysis. RNNs analyze word sequences to comprehend context with the help of attention mechanisms, whereas CNNs extract local patterns; this allows them to zero in on subtleties in hate speech. Findings indicate that deep learning can enhance sentiment analysis in multilingual user-generated content and more accurately detect sentiment in code-mixed data.

Toby (2024) [25] Analyzes social media platforms, such as Reddit, to predict financial market trends, focusing on Bitcoin sentiment from 2010 to 2022, utilizing deep learning and generative AI. This study uses transformers, which combine conventional sentiment analysis techniques with their contextual understanding capabilities, to handle massive text datasets. Generative AI generates synthetic data for rigorous testing and modeling of market sentiment dynamics, while noise-removal techniques ensure that sentiment signals are more discernible. Financial analysts now have cutting-edge resources to understand complicated data better and make better decisions in ever-changing market conditions. The results indicate that social media sentiment can effectively predict market trends.

Chen (2024) [26] utilized data collected from January 2021 to July 2023, sentiment analysis, and deep learning to forecast which terms and words would be widely used on Weibo. The study reveals patterns that impact topic popularity, such as emotional tones in user posts, by incorporating machine learning algorithms with NLP techniques. The language variety in social media is analyzed using feature extraction and preprocessing techniques, such as tokenization and normalization. One way sentiment analysis can foretell changes in public opinion is by monitoring trends through time-series analysis.

E. M. G. Younis, R. Mohamed, A. A. Ali and A. A. Tantawy, (2024) [27] examined Arabic tweets regarding COVID-19 vaccines using a combination of sentiment lexicons and deep learning models. It improves sentiment classification accuracy by combining lexicons that capture subtle emotions and sarcasm with neural networks that recognize patterns. The dataset includes tweets shared at the pandemic's peak, reflecting diverse views on vaccines, ranging from supportive to skeptical. Hybrid approaches effectively address the complexities of multilingual sentiment analysis, and the results demonstrate that sentiment analysis plays a crucial role in informing public health communication strategies.

O. L. Chikaodir, Y. M. Malgwi, S. D. A. Abdulkadir, S. S. Wagbe, F. K. George and W. O. Jatau, [28] suggest combining sentiment analysis with deep learning to identify false news. The model enhances its ability to detect misleading content by integrating word2vec and BERT embeddings, which leverage semantic links and contextual understanding. Sentiment analysis is valuable for spotting emotional undertones that might indicate unfounded assertions. Combining traditional and modern NLP techniques improves feature extraction and system reliability.

TAN (2024) [29] The study demonstrates how crises affect trust in airlines by examining public opinion on incidents such as accidents and service disruptions through social media, news, and government communications. Deep learning improves sentiment analysis by capturing nuanced emotions and mixed sentiments. The study's findings demonstrate the value of real-time insights for crisis communication and trust restoration.

M. Khan and A. Srivastav, (2024) [9] examine how machine learning algorithms can aid sentiment analysis on Twitter. It tests the accuracy of various methods in classifying tweets as positive, negative, or neutral using a dataset from the 2022 FIFA World Cup. These methods include VADER, XGBoost, Random Forest, and LSTM. Bidirectional LSTM outperformed the rest with a 73% accuracy. Tokenization and lemmatization were part of the preprocessing steps.

Alawi & Bozkurt, (2024) [30] A hybrid machine learning model, BERT-BiLSTM-CNN, analyses Turkish university sentiment and satisfaction using Twitter data.

The model addresses the linguistic complexity of Turkish text using BERT transformers, Bidirectional LSTMs, and parallel CNN branches. It has 91% accuracy and robust performance metrics across all evaluation parameters. The hybrid model's ability to analyze public opinion and improve sentiment classification is demonstrated using 17,793 manually annotated tweets.

## III. METHODOLOGY

Sentiment analysis is a multi-stage process [9], starting with data acquisition and moving through preprocessing, model training, and testing phases. This dataset, sourced from the Kaggle repository, contains positive, negative, or neutral sentiment tweets. Contextual data includes user demographics, tweet timing, and other numerical attributes. Data preprocessing included a series of systematic steps to prepare the text data for analysis. This process involved converting text to lowercase, removing special characters and stopwords, and applying tokenization and lemmatization techniques. TF-IDF (Term Frequency-Inverse Document Frequency) vectorization was used to convert processed text data into numerical vectors during feature extraction. Tweet sentiment was classified using machine and deep learning algorithms, such as LSTM, Gradient Boosting, Naive Bayes, Random Forest, and Logistic Regression. We trained and tested these models to classify tweets as positive, negative, or neutral.

### A. Data Collection

This dataset was obtained from the Kaggle Repository. (“https://www.kaggle.com/datasets/abhi8923shriv/sentiment-analysis-dataset”). We used a cross-sectional time horizon, which classifies tweets' sentiments based on data taken at a single point in time [32].

We utilized a historical dataset of tweets that were automatically labeled with emoticons. We labeled emoticons using the assumption that :) Represents positive emotions and :( Represents negative emotions. This dataset has 27482 rows and 10 columns. The dataset (Figure 3) includes the text ID, the tweet, the extracted portion (selected_text), the tweet sentiment, time, age group, country of origin, and numerical data like population, land area, and population density.

### B. Data Pre-processing

Preprocessing was done to clean and standardize the textual dataset for sentiment analysis. Incomplete text field entries were removed. Text data was converted to lowercase, tokenized into individual words, and lemmatized to reduce words to their base forms. Stop words were removed to eliminate irrelevant terms, preserving the dataset's essential features. After processing, the data was split 80:20 into training and testing sets for model evaluation.

### C. Models Implemented

The algorithms that are implemented in this research are described in Table 1.

### A. Implementation and Results of Algorithms

This research employs five unique classifiers.

*Table 1: Implemented Algorithms.*

| | |
|---|---|
| Logistic Regression | It is widely employed to model the relationship between independent variables and two-outcome classification probabilities [16]. |
| Random Forest | This ensemble learning approach integrates the outputs of decision trees to enhance both classification accuracy and robustness [17], [18]. |
| Naive Bayes | This statistical classifier applies Bayes' theorem, assuming that each input feature influences the outcome independently. By estimating the posterior probability for each class and selecting the most likely one, it delivers quick and effective categorisation [18]. |
| Gradient Boosting | In this ensemble approach, a sequence of simple models (often decision trees) is built one after another. Each new model focuses on the mistakes made by its predecessors, steadily reducing residual errors and producing highly accurate predictions [18]. |
| LSTM | It is primarily used for modeling long-term dependencies [19]. |

## IV. RESULTS AND DISCUSSION

The dataset, consisting of 27,482 tweets, was divided into training and testing sets using an 80-20 split, where 80% of the data was allocated for training and 20% for testing. Metrics such as recall, F1-score, and the area under the receiver operating characteristic curve (ROC-AUC) were utilized to evaluate the effectiveness of the selected methods. The equations for these performance metrics are outlined below:

$$\text{Accuracy} = \frac{TP+TN}{TP+TN+FP+FN} \quad (1)$$

The accuracy measure indicates the number of data points accurately projected [33], [34].

$$\text{Precision} = \frac{TP}{TP+FP} \quad (2)$$

The precision metric determines the proportion of predicted positive class samples that are positive [33], [34].

$$\text{Recall} = \frac{TP}{TP+FN} \quad (3)$$

#### 1) Logistic Regression

Recall, also called sensitivity, determines the number of correctly predicted test case samples from all positive classes [33], [34].

$$\text{F1-Score} = 2 * \frac{Precision * Recall}{Precision + Recall} \quad (4)$$

A harmonic mean of recall and precision is the F1-score [33], [34].

The ROC-AUC describes a classifier's overall performance. The values can range from 0 to 1, with one denoting flawless classification and 0.5 denoting incompatible or random classification [33], [34].

The results of these metrics are given in Table 2.

This model achieves a training accuracy of 86.74% and a testing accuracy of 79.56%, indicating a reasonable level of performance.

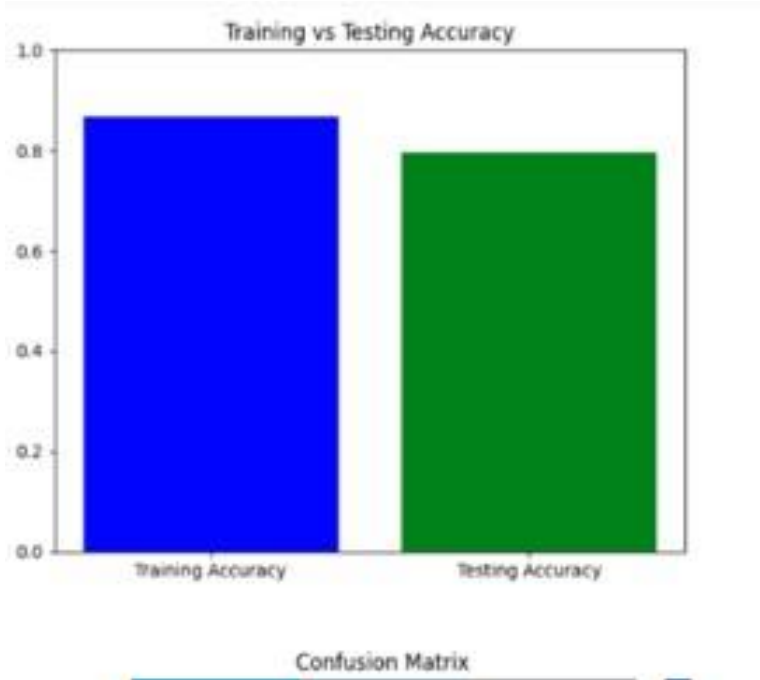

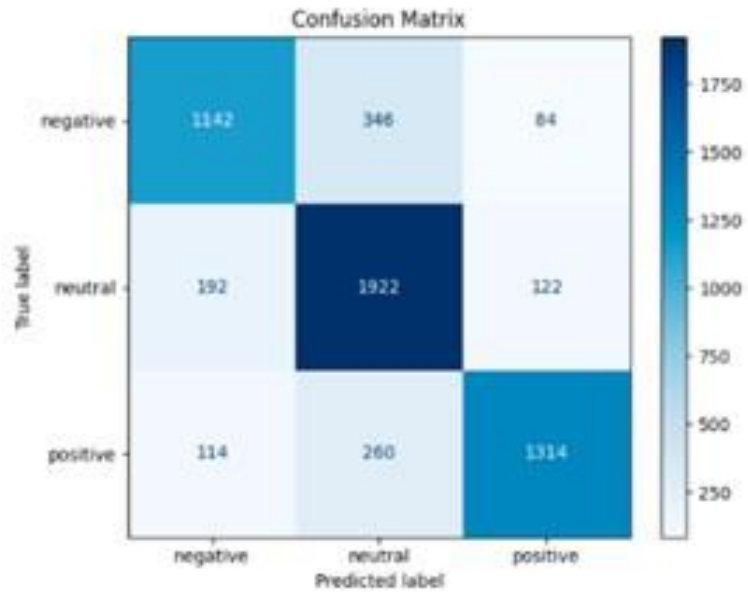

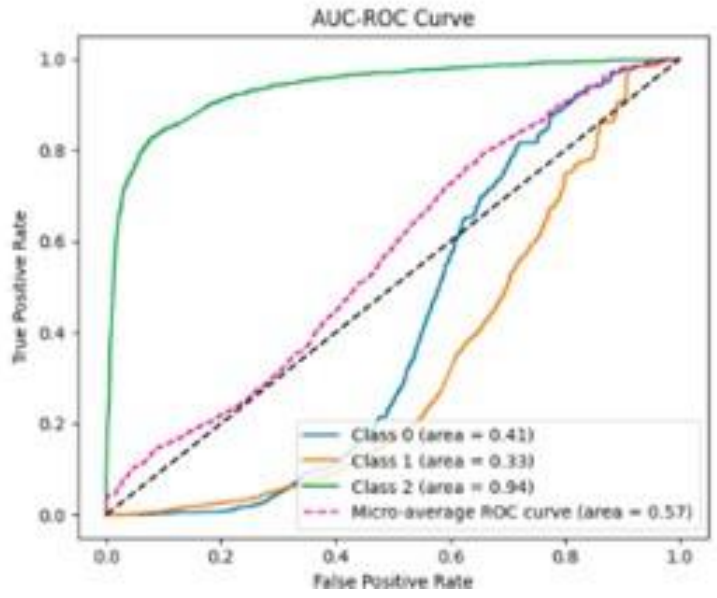

*Figure 1: Performance metrics of the Logistic Regression model.*

The F1 score, recall, and precision are all in the right places, suggesting that the model can reliably predict outcomes across different types of sentiment. The confusion matrix shows that the model sometimes confuses neutral and negative feelings. Class 2 performs well (AUC = 0.94), while Classes 0 and 1 need improvement.

*2) Random Forest*

The model fits the training data well, achieving a training accuracy of 97.70% (Table 2). However, testing accuracy drops to 79.62% (Table 2), suggesting overfitting and inadequate accuracy on unseen data.

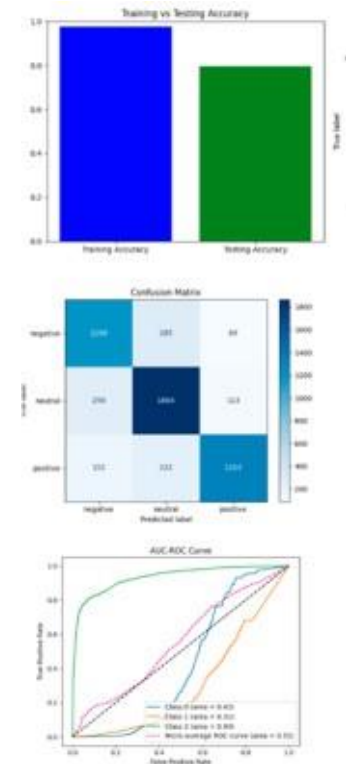

*Figure 2: Performance metrics of the random forest model.*

Like Logistic Regression, it balances precision, recall, and F1 scores around 0.80 (Table 2). The confusion matrix indicates minor positive and negative misclassifications. Class 2 performs well on the AUC-ROC curve (AUC = 0.93) (Table 2), while Classes 0 (neutral) and 1 (negative) are less distinct.

*B. Naïve Bayes*

The Naïve Bayes model achieves 85.90% training accuracy and 78.02% testing accuracy (Table 2), demonstrating reasonable generalization but lagging behind more complex models, such as Random Forests.

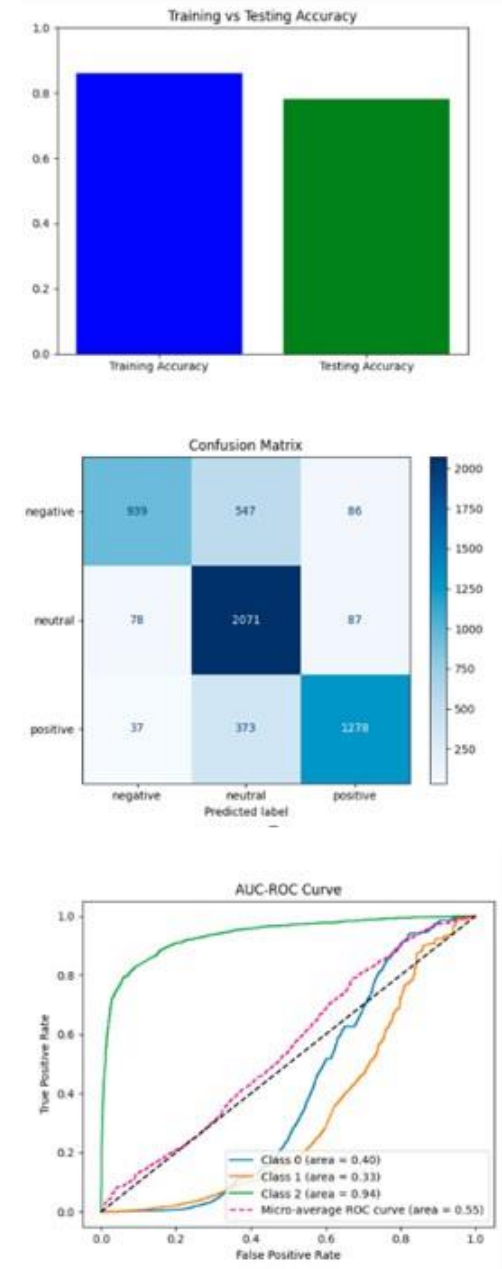

*Figure 3: Performance metrics of the Naïve Bayes model.*

The model performs consistently across classes with recall and precision near 0.80 (Table 2). However, the confusion matrix indicates that distinguishing between positive and negative sentiments is challenging, leading to frequent misclassifications. Class 2 performs well on the AUC-ROC curve (AUC = 0.94) (Table 2), while Classes 0 and 1 struggle to distinguish these sentiments.

*C. Gradient Boosting*

With a training accuracy of 69.89% and a testing accuracy of 67.83% (Table 2), the Gradient Boosting model struggles to generalize to new data.

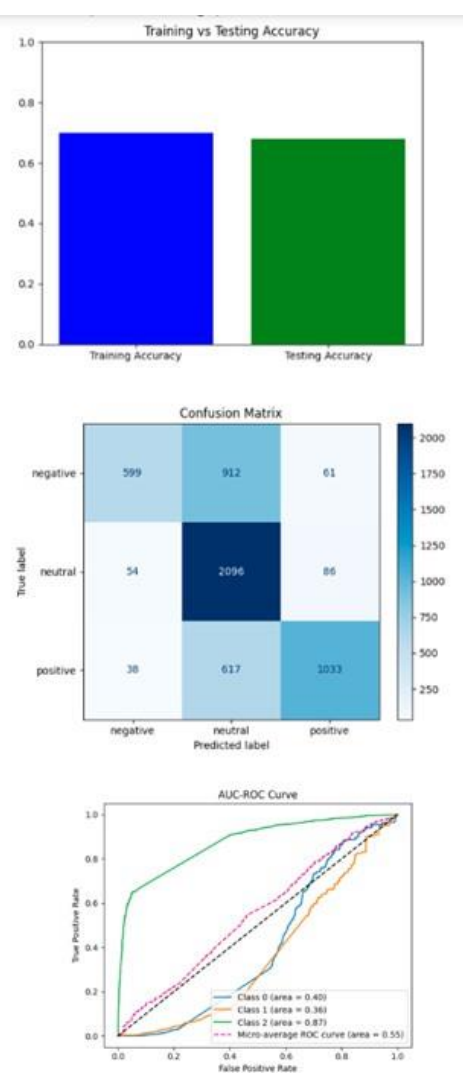

*Figure 4: Performance metrics of the Gradient Boosting model.*

The model exhibits moderate precision (75.21%) (Table 2), but lower recall and F1 scores indicate that it may not be able to identify all critical cases. Many neutral cases are misclassified as negative in the confusion matrix. The AUC-ROC curve shows that Class 2 performs well, but Classes 0 and 1 struggle, resulting in suboptimal model performance.

### *D. LSTM*

The LSTM model achieves 90.98% training accuracy and 80.00% testing accuracy (Table 2), indicating good generalization to unseen data. Precision, recall, and F1 scores around 0.80 indicate balanced performance across sentiment classes.

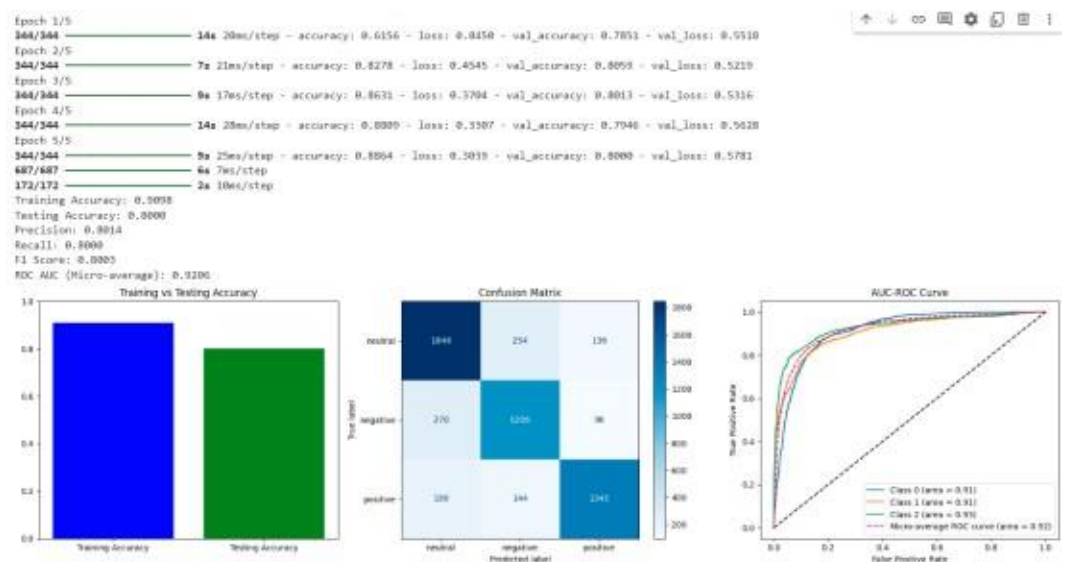

*Figure 5: Performance metrics of the LSTM model.*

The confusion matrix shows that the model detects positive and neutral sentiments but misclassifies negative sentiments. AUC-ROC values above 0.91 for all classes and a micro-average AUC of 0.92 demonstrate the model's ability to distinguish sentiment categories. These outcomes demonstrate the LSTM model's ability to handle the complexity of sentiment classification and provide reliable predictions.

*Table 2: Performance Results of Algorithms.*

| Model | Logistic Regression | Random Forest | Naive Bayes | Gradient Boosting | LSTM |
|---|---|---|---|---|---|
| Training Accuracy | 0.8674 | 0.977 | 0.859 | 0.6989 | 0.9098 |
| Testing Accuracy | 0.7966 | 0.7962 | 0.7802 | 0.6783 | 0.8 |
| Precision | 0.8004 | 0.799 | 0.807 | 0.7521 | 0.8014 |
| Recall | 0.7966 | 0.7962 | 0.7802 | 0.6783 | 0.8 |
| F1 Score | 0.7962 | 0.7966 | 0.777 | 0.6637 | 0.8003 |
| ROC AUC (Micro-average) | 0.5668 | 0.5535 | 0.5498 | 0.5484 | 0.9206 |

## V. CONCLUSION

This paper examines the performance of five complex machine learning and deep learning approaches, Logistic Regression, Random Forest, Naïve Bayes, Gradient Boost, and LSTM, on sentiment analysis of Twitter data. The LSTM model emerges as the most effective, achieving a training accuracy of 90.98%, testing accuracy of 80.00%, balanced precision, recall, and F1 scores (~0.80), and a micro-average ROC-AUC of 0.92. These results highlight the LSTM model's ability to capture contextual and sequential nuances in textual data, outperforming traditional machine learning models. This research utilizes a Kaggle dataset that has been vectorized, tokenized, lemmatized, and stopwords removed to prepare it for analysis. The significance of thorough preprocessing and model assessment is highlighted in this study.

Further work is to create a stronger customer insights platform, integrate sentiment analysis with other predictive models, such as those for customer churn or product recommendations.

## VI. References

[1] N. Y. Hussain, "Deep Learning Architectures Enabling Sophisticated Feature Extraction and Representation for Complex Data Analysis," International Journal of Innovative Science and Research Technology, vol. 9, no. 10, pp. 2290-2300, 2024.

[2] S. Garcia, F. Gomez-Donoso and M. Cazorla, "Enhancing Human–Robot Interaction: Development of Multimodal Robotic Assistant for User Emotion Recognition," New Insights into Rehabilitation Robots with Intelligent Sensing Systems, vol. 14, no. 24, p. 11914, 2024.

[3] S. Lai, X. Hu, H. Xu, Z. Ren and Z. Liu, "Multimodal Sentiment Analysis: A Survey," Displays, vol. 80, pp. 1-17, 2023.

[4] D. Kanojia and A. Joshi, "Applications and Challenges of Sentiment Analysis in Real-life Scenarios," Computation and Language, 2023.

[5] K. L. Tan, C. P. Lee and K. M. Lim, "A Survey of Sentiment Analysis: Approaches, Datasets, and Future Research," Applied Sciences, vol. 13, no. 7, pp. 1-21, 2023.

[6] D. Dangi, D. K. Dixit and A. Bhagat, "Sentiment analysis of COVID-19 social media data through machine learning," Multimedia Tools and Applications, vol. 81, p. 42261–42283, 2022.

[7] A. C. M. V. Srinivas, Ch.Satyanarayana, Ch.Divakar and K. P. Sirisha, "Sentiment Analysis using Neural Network and LSTM," IOP Conference Series: Materials Science and Engineering, vol. 1074, no. 1, pp. 1-7, 2021.

[8] M. S. Islam, M. N. Kabir, N. A. Ghani, K. Z. Zamli, N. S. A. Zulkifli, M. M. Rahman and M. A. Moni, "Challenges and future in deep learning for sentiment analysis: a comprehensive review and a proposed novel hybrid approach," Artificial Intelligence Review, vol. 57, no. 62, pp. 1-79, 2024.

[9] M. Khan and A. Srivastav, "Sentiment Analysis of Twitter Data Using Machine Learning Techniques," International Journal of Engineering and Management Research, vol. 4, no. 1, pp. 196-203, 2024.

[10] K. Gulatia, S. S. Kumarb, R. S. K. Bodduc, K. Sarvakard, D. K. Sharmae and M. Noman, "Comparative analysis of machine learning-based classification modelsusing sentiment classification of tweets related to COVID-19 pandemic," Materials Today: Proceeding, vol. 51, no. 1, pp. 38-41, 2022.

[11] S. K. Bharti, Varadhaganapathy, R. K. Gupta, P. K. Shukla, M. Bouye, S. K. Hingaa and A. Mahmoud, "Text-Based Emotion Recognition Using Deep Learning Approach," Computational Intelligence and Neuroscience, vol. 45, no. 60, pp. 1-8, 2022.

[12] T. Diwan and J. V. Tembhurne, "Sentiment analysis: a convolutional neural networks perspective.," Multimedia Tools and Applications, vol. 81, no. 6, p. 44405–44429, 2022.

[13] Q. Liu, Z. Tao, Y. Tse and C. Wang, "Stock market prediction with deep learning: the case of china. Finance," Finance Research Letters, vol. 46, pp. 1-10, 2022.

[14] A. Mewada and R. K. DEWANG, "Sa-asba: a hybrid model for aspect-based sentiment analysis using synthetic attention in pre-trained language bert model with extreme gradient boosting.," The Journal of Supercomputing , vol. 79, no. 3, pp. 5516-5551, 2023.

[15] P. D. Mahendhiran and K. Subramanian, "Deep learning techniques for polarity classifcation in multimodal Sentiment Analysis," International Journal of Information Technology & Decision Making, vol. 17, no. 1, pp. 883-910, 2018.

[16] J. He, R. N. Makarov, J. Tuero and Z. Wang, "Performance evaluation metric for statistical learning trading strategies," Data Science in Finance and Economics, vol. 4, no. 4, pp. 570-600, 2024.

[17] Z. Sun, G. Wang, P. Li, H. Wang, M. Zhang and X. Liang, "An improved random forest based on the classification accuracy and correlation measurement of decision trees," Expert Systems with Applications, vol. 237, 2024.

[18] V. Nakhipova, Y. Kerimbekov, Z. Umarova, L. Suleimenova, S. Botayeva, A. Ibashova and N. Zhumatayev, "Use of the Naive Bayes Classifier Algorithm in Machine Learning for Student Performance Prediction," International Journal of Information and Education Technology, vol. 14, no. 1, pp. 92-98, 2024.

[19] B. Lindemann, T. Müller, H. Vietz, N. Jazdi and M. Weyrich, "A survey on long short-term memory networks for time series prediction," 14th CIRP Conference on Intelligent Computation in Manufacturing Engineering, pp. 650-655, 2021.

[20] D. Narayanan, S. P. Jaganathan and D. Thiyagarajan, "Sentimental Analysis Recognition in Customer review using Novel-CNN," Conference: 2023 International Conference on Computer Communication and Informatics (ICCCI), 2023.

[21] Y. Li, "Optimizing Sentiment Analysis of User Reviews and Emotional Marketing Strategies on E-commerce Platforms Using Deep Learning Algorithms," Journal of Electrical Systems , pp. 437-444, 2024.

[22] G. Srivastav, S. Kant, D. Srivastava, N. Sharma and Y.-C. Hu, "An efficient sentiment analysis technique based on fine-tuned EdBERT for virtual learning environments," Multimedia Tools and Applications, 2024.

[23] R. Geethanjali and A. VALARMATHI, "A Novel Hybrid Deep Learning IChOA-CNN-LSTM Model for Modality-Enriched and Multilingual Emotion Recognition in Social Media A Novel Hybrid Deep Learning IChOA-CNN-LSTM Model," Research Square, vol. 14, no. 1, 2024.

[24] S. Chanda and S. Pal, "Hate Content Identification in Code-mixed Social Media Data," Text and Social Media Analytics for Fake News and Hate Speech Detection, 2024.

[25] W. J. Toby, "Transformers and tradition: using Generative AI and Deep Learning for financial markets prediction," London School of Economics and Political Science, 2024.

[26] Y. Chen, "Analysis and prediction of factors influencing the hotness of Weibo trending keywords," Applied and Computational Engineering, vol. 55, no. 1, pp. 110-119, 2024.

[27] E. M. G. Younis, R. Mohamed, A. A. Ali and A. A. Tantawy, "Hybrid annotation and classification for predicting attitudes towards COVID-19 vaccines for Arabic tweets," Social Network Analysis and Mining, vol. 14, no. 137, 2024.

[28] O. L. Chikaodir, Y. M. Malgwi, S. D. A. Abdulkadir, S. S. Wagbe, F. K. George and W. O. Jatau, "Enhanced Fake News Detection Model using Hybridization of word2vec and BERT Embedding's," International Journal of Current Researches in Sciences, Social Sciences and Languages (IJCRSSSL), pp. 63-76, 2024.

[29] K. Y. TAN, "Learning to outsmart a crisis: A strategic management view on managing aviation crises with machine learning," International Conference on Management and Business, pp. 1-132, 2024.

[30] A. B. Alawi and F. Bozkurt, "A hybrid machine learning model for sentiment analysis and satisfaction," Decision Analytics Journal, vol. 11, pp. 1-13, 2024.

[31] M. M. Rahman, A. I. Shiplu, Y. Watanobe and M. A. Alam, "RoBERTa-BiLSTM: A Context-Aware Hybrid Model for Sentiment Analysis," JOURNAL OF LATEX CLASS FILES, vol. 14, no. 8, pp. 1-18, 2024.

[32] M. Islam and S. Jin, "An Overview of Data Visualization," in International Conference on Information Science and Communications Technologies (ICISCT), Tashkent, Uzbekistan, 2019.

[33] M. Padhy, U. M. Modibbo, R. Rautray, S. S. Tripathy and S. Bebortta, "Application of Machine Learning Techniques to Classify Twitter Sentiments Using Vectorization Techniques," Algorithms, vol. 17, no. 11, p. 486, 2024.

[34] M. Kuang, R. Safa, S. A. Edalatpanah and R. S. Keyser, " A HYBRID DEEP LEARNING APPROACH FOR SENTIMENT ANALYSIS IN PRODUCT REVIEWS," Mechanical Engineering , vol. 21, no. 3, pp. 479-500, 2023.